\documentclass[sigconf]{acmart}

\AtBeginDocument{%
  \providecommand\BibTeX{{%
    \normalfont B\kern-0.5em{\scshape i\kern-0.25em b}\kern-0.8em\TeX}}}

\acmDOI{}

\acmConference{}{}{}{}
\acmBooktitle{}
\acmPrice{15.00}
\acmISBN{978-1-4503-XXXX-X/18/06}

\usepackage{amsmath}
\usepackage{mathtools}
\usepackage{amsthm}

\usepackage{graphicx}
\usepackage{hyperref}
\usepackage[capitalize,noabbrev]{cleveref}

\usepackage{booktabs} 
\usepackage{xcolor}   
\usepackage{color}
\usepackage{subcaption}
\usepackage{caption}
\usepackage{comment}
\usepackage{multirow,setspace}
\usepackage{enumitem}
\usepackage{titlesec}
\usepackage{relsize}
\usepackage[normalem]{ulem}
\usepackage{xfrac}
\usepackage{nicefrac}       
\usepackage{etoolbox}
\usepackage{microtype}      
\usepackage{adjustbox}         
\usepackage{multirow,siunitx}
\usepackage{comment}
\usepackage{stmaryrd}
\usepackage{diagbox}
\usepackage{algorithm}
\usepackage{algorithmicx}
\usepackage[noend]{algpseudocode}

\usepackage{pgf}
\usepackage{tikz}

\usetikzlibrary{arrows,decorations.pathmorphing,decorations.footprints,fadings,calc,trees,mindmap,shadows,decorations.text,patterns,positioning,shapes,matrix,fit}
\usetikzlibrary{arrows,shadows,backgrounds}
\usetikzlibrary{arrows.meta}
\usetikzlibrary{positioning,fit}
\usetikzlibrary{automata}
\usetikzlibrary{matrix}
\usetikzlibrary{shapes.symbols,shapes.misc,shapes.arrows}
\usetikzlibrary{matrix,chains,scopes,decorations.pathmorphing}
\usetikzlibrary{bayesnet}

\usepackage[textsize=tiny]{todonotes}

\theoremstyle{plain}
\newtheorem{theorem}{Theorem}[section]

\theoremstyle{definition}
\newtheorem{definition}[theorem]{Definition}

\theoremstyle{remark}

\crefname{theorem}{Theorem}{Theorems}
\crefname{lemma}{Lemma}{Lemmas}
\crefname{proposition}{Proposition}{Propositions}
\crefname{definition}{Definition}{Definitions}
\crefname{corollary}{Corollary}{Corollaries}
\crefname{example}{Example}{Examples}
\crefname{claim}{Claim}{Claims}
\crefname{assumption}{Assumption}{Assumptions}

\newcommand{\fml}[1]{{\mathcal{#1}}}

\newcommand{\tn}[1]{\textnormal{#1}}

\newcommand{\mbf}[1]{\ensuremath\mathbf{#1}}
\newcommand{\msf}[1]{\ensuremath\mathsf{#1}}
\newcommand{\mbb}[1]{\ensuremath\mathbb{#1}}

\newcommand{\tsf}[1]{{\textsf{\small #1}}}

\newcommand{\Prob}{\ensuremath\tn{Pr}}
\newcommand{\prob}{\Prob}

\newcommand{\waxp}{\ensuremath\mathsf{WeakAXp}}

\newcommand{\axp}{\ensuremath\mathsf{AXp}}

\newcommand{\isweakpaxp}{\ensuremath\mathsf{isWeakPAXp}}
\newcommand{\wdrset}{\ensuremath\mathsf{WeakPAXp}}
\newcommand{\drset}{\ensuremath\mathsf{PAXp}}
\newcommand{\mdrset}{\ensuremath\mathsf{MinPAXp}}
\newcommand{\adrset}{\ensuremath\mathsf{ApproxPAXp}}

\DeclareMathOperator*{\limply}{\rightarrow}

\DeclareMathOperator*{\argmax}{\msf{argmax}}
\DeclareMathOperator*{\lprob}{lPr}

\newcommand{\lvt}{\mathbf{t}}
\newcommand{\lvf}{\mathbf{f}}

\definecolor{gray}{rgb}{.4,.4,.4}
\definecolor{midgrey}{rgb}{0.5,0.5,0.5}
\definecolor{middarkgrey}{rgb}{0.35,0.35,0.35}
\definecolor{darkgrey}{rgb}{0.3,0.3,0.3}
\definecolor{darkred}{rgb}{0.7,0.1,0.1}
\definecolor{midblue}{rgb}{0.2,0.2,0.7}
\definecolor{darkblue}{rgb}{0.1,0.1,0.5}
\definecolor{darkgreen}{rgb}{0.1,0.5,0.1}
\definecolor{defseagreen}{cmyk}{0.69,0,0.50,0}

\newcommand{\jnoteF}[1]{}

\newcolumntype{L}[1]{>{\raggedright\let\newline\\\arraybackslash\hspace{0pt}}m{#1}}
\newcolumntype{C}[1]{>{\centering\let\newline\\\arraybackslash\hspace{0pt}}m{#1}}
\newcolumntype{R}[1]{>{\raggedleft\let\newline\\\arraybackslash\hspace{0pt}}m{#1}}

\tikzset{
  0 my edge/.style={densely dashed, my edge},
  my edge/.style={-{Stealth[]}},
}

\setlist{nosep}

\providetoggle{long}
\settoggle{long}{false}

\begin{document}

\title{On Computing Relevant Features for Explaining NBCs}


\author{Yacine Izza}
\affiliation{%
  \institution{University of Toulouse}
  \city{Toulouse}
  \country{France}}
\email{yacine.izza@univ-toulouse.fr}

\author{Joao Marques-Silva}
\affiliation{%
  \institution{IRIT, CNRS}
  \city{Toulouse}
  \country{France}}
\email{joao.marques-silva@irit.fr}

\renewcommand{\shortauthors}{Izza and Marques-Silva}




\begin{abstract}
  Despite the progress observed with model-agnostic explainable AI
  (XAI), it is the case that model-agnostic XAI can produce incorrect 
  explanations.
  One alternative are the so-called formal approaches to XAI,
  that include PI-explanations.  Unfortunately, PI-explanations also 
  exhibit important drawbacks, the most visible of which is arguably
  their size.  
  %
  The computation of relevant features serves to trade off
  probabilistic precision for the number of features in an
  explanation.
  However, even for very simple classifiers, the complexity of
  computing sets of relevant features
  is prohibitive.
  This paper investigates the computation of
  relevant sets 
  for Naive Bayes Classifiers (NBCs), and shows that, in practice,
  these are easy to compute. Furthermore, the experiments confirm that
  succinct sets of relevant features
  can be obtained with NBCs.
\end{abstract}

\maketitle

\section{Introduction} \label{sec:intro}

The advances in Machine Learning (ML) in recent years motivate an ever
increasing range of practical applications of Artificial Intelligence
(AI) systems. In some domains, the use of AI systems is premised on
the availability of mechanisms for explaining the often opaque
operation of ML models.
Some uses of ML models are deemed \emph{high-risk} given the impact
that their operation can have on people~\cite{eu-aiact21}. (Other
authors refer to \emph{high-stakes}
applications~\cite{rudin-naturemi19}.)
For high-risk AI systems, a critical requirement is rigor, either when
reasoning about these systems, or when explaining their
predictions.

Recent years have witnessed a growing interest in
%
%
eXplainable AI (XAI)~\cite{muller-dsp18,pedreschi-acmcs19,xai-bk19,muller-xai19-ch01,molnar-bk20,muller-ieee-proc21}. The best-known XAI
approaches can be broadly categorized as model-agnostic methods, that
include for example LIME~\cite{guestrin-kdd16},
SHAP~\cite{lundberg-nips17} and Anchor~\cite{guestrin-aaai18}, and
intrinsic interpretability~\cite{rudin-naturemi19,molnar-bk20}, for
which the explanation is represented by the actual (interpretable) ML
model.
Intrinsic interpretability may not represent a viable option in some
uses of AI systems. On the other hand, model-agnostic methods, although
locally accurate, can produce explanations that are
unsound~\cite{ignatiev-ijcai20}, in addition to displaying several
other
drawbacks~\cite{lukasiewicz-corr19,lakkaraju-aies20a,lakkaraju-aies20b,weller-ecai20}.
Unsound explanations are hopeless whenever rigor is a key requirement;
thus, model-agnostic explanations ought not be used  in high-risk
settings.
%
Indeed, it has been reported~\cite{ignatiev-ijcai20} that an
explanation $X$ can be consistent with different predicted classes.
For example, for a bank loan application, $X$ might be consistent with 
an approved loan application, but also with a declined loan
application.
An explanation that is consistent with both a declined and an approved
loan applications offers no insight to why one of the loan
applications was declined.
%
%
There have been recent efforts on rigorous XAI
approaches~\cite{darwiche-ijcai18,inms-aaai19,inms-nips19,nsmims-sat19,iims-corr20,inams-aiia20,darwiche-ecai20,marquis-kr20,ims-ijcai21,hiims-kr21,cms-cp21,imsns-ijcai21,msgcin-icml21,ims-sat21,kwiatkowska-ijcai21,mazure-cikm21,rubin-aaai22,hiicams-aaai22,snimmsv-aaai22,iisms-aaai22,msi-aaai22,iims-corr22a,ignatiev-corr22},
most of which are based on feature attribution, namely the computation
of so-called abductive explanations (AXp's). However, these efforts
have  mostly focused on the scalability of computing rigorous
explanations, with more recent work investigating input
distributions~\cite{rubin-aaai22}. Nevertheless, another important
limitation of rigorous XAI approaches is the often unwieldy size of
explanations.
Recent work studied probabilistic explanations, as a mechanism to
reduce the size of rigorous
explanations~\cite{vandenbroeck-ijcai21,kutyniok-jair21}.
Probabilistic explanations have extended model-agnostic
approaches~\cite{vandenbroeck-ijcai21}, and so can suffer from 
unsoundness. Alternatively, more rigorous approaches to computing
probabilistic explanations have been shown to be computationally hard,
concretely hard for $\tn{NP}^{PP}$, and so most likely beyond the
reach of modern automated reasoners.

This paper builds on recent  work~\cite{kutyniok-jair21} on rigorous
probabilistic explanations, and investigates their practical
scalability.
However, instead of considering classifiers represented as boolean
circuits (as in~\cite{kutyniok-jair21}), the paper specifically
considers the family of naive Bayes classifiers (NBCs). 
Earlier work showed that rigorous explanations of NBCs, concretely
AXp's, are computed in polynomial time, and that their enumeration
is achieved with polynomial delay~\cite{msgcin-nips20}.
Unfortunately, the size of explanations was not investigated in this
earlier work. 
This paper studies probabilistic explanations for the concrete case of
NBCs. For the case of categorical features, the paper relates
probabilistic explanations of NBCs with the problem of counting the
models of (restricted forms) of integer programming constraints, and
proposes a dynamic programming based, pseudo-polynomial algorithm for
computing approximate explanations.
Such approximate explanations offer important theoretical guarantees:
i) approximate explanations are not larger than some rigorous
explanation; ii) approximate explanations are not smaller than some
rigorous probabilistic explanation; and iii) approximate explanations
offer strong probabilistic guarantees on their precision.
More importantly, the experimental results demonstrate that succinct
explanations, with sizes that can be deemed within the grasp of human
decision makers~\cite{miller-pr56}, can be very efficiently computed
with most often a small decrease in the precision of the explanation. 

The paper is organized as follows.
\cref{sec:prelim} introduces the definitions and notation used
throughout the paper.
\cref{sec:xlc} summarizes the computation of explanations for NBCs
proposed in earlier work~\cite{msgcin-nips20}.
\cref{sec:paxp} details the approach proposed in this paper for
computing approximate probabilistic AXp's.
%
%
\cref{sec:res} presents experimental results confirming that precise
short approximate AXp's can be efficiently computed.
\cref{sec:conc} concludes the paper.

\section{Preliminaries} \label{sec:prelim}

\subsection{Classification problems}
This paper considers classification problems, which are defined on a
set of features (or attributes) $\fml{F}=\{1,\ldots,m\}$ and a set of
classes $\fml{K}=\{c_1,c_2,\ldots,c_K\}$.
Each feature $i\in\fml{F}$ takes values from a domain $\mbb{D}_i$.
In general, domains can be categorical or ordinal, with values that
can be boolean, integer or real-valued
but in this paper we restrict  $\fml{K}=\{0,1\}$, i.e.\ binary
classifiers, and all features are categorical.  
(Throughout the paper, we also use the notations $\ominus$ and 
$\oplus$ to denote, resp.\ class 0 and class 1.) 
Feature space is defined as
$\mbb{F}=\mbb{D}_1\times{\mbb{D}_2}\times\ldots\times{\mbb{D}_m}$;
$|\mbb{F}|$ represents the total number of points in $\mbb{F}$ if none
of the features is real-valued.
For boolean domains, $\mbb{D}_i=\{0,1\}=\mbb{B}$, $i=1,\ldots,m$, and
$\mbb{F}=\mbb{B}^{m}$.
The notation $\mbf{x}=(x_1,\ldots,x_m)$ denotes an arbitrary point in
feature space, where each $x_i$ is a variable taking values from
$\mbb{D}_i$. The set of variables associated with features is
$X=\{x_1,\ldots,x_m\}$.
Moreover, the notation $\mbf{v}=(v_1,\ldots,v_m)$ represents a
specific point in feature space, where each $v_i$ is a constant
representing one concrete value from $\mbb{D}_i$. 
An ML classifier $\mbb{M}$ is characterized by a (non-constant)
\emph{classification function} $\kappa$ that maps feature space
$\mbb{F}$ into the set of classes $\fml{K}$,
i.e.\ $\kappa:\mbb{F}\to\fml{K}$.
An \emph{instance} (or observation)
denotes a pair $(\mbf{v}, c)$, where $\mbf{v}\in\mbb{F}$ and
$c\in\fml{K}$, with $c=\kappa(\mbf{v})$. 
(We also use the term \emph{instance} to refer to $\mbf{v}$, leaving
$c$ implicit.)

\subsection{Formal explanations}
We now define formal explanations. In contrast with the well-known
model-agnostic approaches to
XAI~\cite{guestrin-kdd16,lundberg-nips17,guestrin-aaai18,pedreschi-acmcs19},
formal explanations are model-precise, i.e.\ their definition reflects
the model's computed function.
Prime implicant (PI) explanations~\cite{darwiche-ijcai18} denote a
minimal set of literals (relating a feature value $x_i$ and a constant
$v_i\in\mbb{D}_i$) 
that are sufficient for the prediction. PI-explanations are related
with abduction, and so are also referred to as abductive explanations
($\axp$)~\cite{inms-aaai19}.
Formally, given $\mbf{v}=(v_1,\ldots,v_m)\in\mbb{F}$ with
$\kappa(\mbf{v})=c$, an $\axp$ is any minimal subset
$\fml{X}\subseteq\fml{F}$ such that, 
\begin{equation} \label{eq:axp}
  \forall(\mbf{x}\in\mbb{F}).
  \left[
    \bigwedge\nolimits_{i\in{\fml{X}}}(x_i=v_i)
    \right]
  \limply(\kappa(\mbf{x})=c)
\end{equation}
i.e.\ the features in $\fml{X}$ are sufficient for the prediction
when these take the values dictated by $\mbf{v}$, and $\fml{X}$ is
irreducible. Also, a non-minimal set such that ~\eqref{eq:axp} holds
is a $\waxp$.
%
$\axp$'s can be viewed as answering a `Why?' question, i.e.\ why is some 
prediction made given some point in feature space.
%
%
Contrastive explanations~\cite{miller-aij19} offer a different view of
explanations, but these are beyond the scope of the paper.
%
%


\subsection{$\delta$-relevant sets}
$\delta$-relevant sets were proposed in more recent
work~\cite{kutyniok-jair21} as a generalized formalization of 
explanations. $\delta$-relevant sets can be viewed as
\emph{probabilistic} PIs,
with $\axp$'s representing a special case of $\delta$-relevant 
sets where $\delta = 1$,
i.e.\ probabilistic PIs that are actual PIs. We briefly overview the
definitions related with relevant sets. 
%
%
The assumptions regarding the probabilities of logical propositions
are those made in earlier work~\cite{kutyniok-jair21}.
Let $\prob_{\mbf{x}}(A(\mbf{x}))$ denote the probability of some
proposition $A$ defined on the vector of variables
$\mbf{x}=(x_1,\ldots,x_m)$, i.e.
\begin{equation} \label{eq:pdefs}
  \begin{array}{rcl}
    \prob_{\mbf{x}}(A(\mbf{x})) & = &
    \frac{|\{\mbf{x}\in\mbb{F}:A(\mbf{x})=1\}|}{|\{\mbf{x}\in\mbb{F}\}|}
    \\[8.0pt]
    \prob_{\mbf{x}}(A(\mbf{x})\,|\,B(\mbf{x})) & = &
    \frac{|\{\mbf{x}\in\mbb{F}:A(\mbf{x})=1\land{B(\mbf{x})=1}\}|}{|\{\mbf{x}\in\mbb{F}:B(\mbf{x})=1\}|}
  \end{array}
\end{equation}
(Similar to earlier work, it is assumed that the features are
independent and uniformly distributed~\cite{kutyniok-jair21}.
Moreover, the definitions above can be adapted in case some of the
features are real-valued. As noted earlier, 
the paper studies only categorical features.) 

\begin{definition}[$\delta$-relevant set~\cite{kutyniok-jair21}]\label{def:drs}
  Consider $\kappa:\mbb{B}^{m}\to\fml{K}=\mbb{B}$, $\mbf{v}\in\mbb{B}^m$,
  $\kappa(\mbf{v})=c\in\mbb{B}$, and
  $\delta\in[0,1]$. $\fml{S}\subseteq\fml{F}$ is a $\delta$-relevant
  set for $\kappa$ and $\mbf{v}$ if,
  \begin{equation} \label{eq:drs}
    \prob_{\mbf{x}}(\kappa(\mbf{x})=c\,|\,\mbf{x}_{\fml{S}}=\mbf{v}_{\fml{S}})\ge\delta
  \end{equation}
  (where the restriction of $\mbf{x}$ to the variables with indices in
  $\fml{S}$ is represented by
  $\mbf{x}_{\fml{S}}=(x_i)_{i\in\fml{S}}$).
\end{definition}
(Observe that
$\prob_{\mbf{x}}(\kappa(\mbf{x})=c\,|\,\mbf{x}_{\fml{S}}=\mbf{v}_{\fml{S}})$ 
is often referred to as the \emph{precision} of
$\fml{S}$~\cite{guestrin-aaai18,nsmims-sat19}.)
Thus, a $\delta$-relevant set represents a set of features which, if
fixed to some pre-defined value (taken from a reference vector
$\mbf{v}$), ensures that the probability of the prediction being the
same as the one for $\mbf{v}$ is no less than $\delta$. 

\begin{definition}[Min-$\delta$-relevant set] \label{def:mdrs}
  Given $\kappa$, $\mbf{v}\in\mbb{B}^{m}$, and $\delta\in[0,1]$, find
  the smallest $k$, such that there exists $\fml{S}\subseteq\fml{F}$, with
  $|\fml{S}|={k}$, and $\fml{S}$ is a $\delta$-relevant set for
  $\kappa$ and $\mbf{v}$.
\end{definition}
With the goal of proving the computational complexity of finding a
minimum-size set of features that is a $\delta$-relevant set, earlier
work~\cite{kutyniok-jair21} restricted the definition to the case
where $\kappa$ is represented as a boolean circuit.

  (Boolean circuits were restricted to propositional formulas defined 
  using the operators $\lor$, $\land$ and $\neg$, and using a set of
  variables representing the inputs; this explains the choice of
  \emph{inputs} over \emph{sets} in earlier
  work~\cite{kutyniok-jair21}.)%

\subsection{Naive Bayes Classifiers (NBCs)}
NBC~\cite{duda-bk73} is a Bayesian Network model~\cite{friedman-ml97} 
characterized by  strong conditional independence  assumptions 
among the features.
%
%
%
Given some observation $\mbf{x} \in \mbb{F}$, the predicted class 
is given by:
\begin{equation} \label{eq:nbc1}
  \kappa(\mbf{x}) = \argmax\nolimits_{c\in\fml{K}}\left(\prob(c|\mbf{x})\right)
\end{equation}
Using the Bayes theorem, $\prob(c|\mbf{x})$ can be computed as follows:
$\prob(c|\mbf{x})=\nicefrac{\prob(c,\mbf{x})}{\prob(\mbf{x})}$.
In practice, we compute only the numerator of the fraction, since 
the denominator $\prob(\mbf{x})$ is constant for every $c\in\fml{K}$. 
Moreover, given the conditional mutual independency of the features, 
we have:  
\[\prob(c,\mbf{x}) = \prob(c)\times\prod\nolimits_i\prob(x_i|c)\] 
Furthermore, it is also common in practice to apply logarithmic 
transformations on probabilities of $\prob(c,\mbf{x})$, thus getting:
\[ \log \prob(c,\mbf{x}) = \log{\prob(c)}+\sum\nolimits_i\log{\prob(x_i|c)} \]
Therefore,~\eqref{eq:nbc1} can be rewritten as follows: 
\begin{equation} \label{eq:nbc4}
  \kappa(\mbf{x}) =
  \argmax\nolimits_{c\in\fml{K}}\left(\log{\prob(c)}+\sum\nolimits_i\log{\prob(x_i|c)}\right)
\end{equation}
For simplicity, and following the notations used in  \cite{msgcin-nips20}, 
we use $\lprob$ to denote the logarithmic probabilities, thus getting:
\begin{equation} \label{eq:nbc5}
  \kappa(\mbf{x}) =
  \argmax\nolimits_{c\in\fml{K}}\left(\lprob(c)+\sum\nolimits_i\lprob(x_i|c)\right)
\end{equation}

(Note that also for simplicity, it is common in  practice to add 
a sufficiently large  positive threshold $T$ to each probability and 
then use only positive values.)
%

\begin{figure*}[t]
  \begin{center}
    \scalebox{0.9125}{
\begin{tikzpicture} 
  \node[latent]                     (G)      {$G$};   %
  \node[latent,below left=1.75cm and 2.0cm of G]    (R2)     {$R_2$}; %
   \node[latent,right=1.75cm of R2]   (R3)     {$R_3$}; %
  \node[latent,left=1.75cm of R2]    (R1)     {$R_1$}; %
  \node[latent,below right=1.75cm and 2.0cm of G]   (R4)     {$R_4$}; %
  \node[latent,right=1.75cm of R4]  (R5)     {$R_5$}; %

  \edge[->] {G} {R1,R2,R3,R4,R5} ;


  \node[right=0.5cm of G,yshift=5pt] (CPT0) { 
    \begin{tabular}{|c|c|}\hline
      $G$ & $\prob(G)$ \\ \hline
      $\ominus$ & 0.90 \\ \hline
  \end{tabular} } ;
  
  \node[left=0.5cm of R1,yshift=-10pt] (CPT1) {
    \begin{tabular}{|c|c|}\hline
      $G$ & $\prob(R_1|G)$ \\ \hline
      $\oplus$ & 0.95 \\\hline
      $\ominus$ & 0.03 \\\hline\end{tabular} } ;
  
  \node[below=0.35cm of R2,xshift=-10pt] (CPT2) {
    \begin{tabular}{|c|c|}\hline
      $G$ & $\prob(R_2|G)$ \\ \hline
      $\oplus$ & 0.05 \\\hline
      $\ominus$ & 0.95 \\\hline\end{tabular} } ; 
  
  \node[below=0.35cm of R3,xshift=0pt] (CPT3) {
    \begin{tabular}{|c|c|}\hline
      $G$ & $\prob(R_3|G)$ \\ \hline
      $\oplus$ & 0.02 \\\hline
      $\ominus$ & 0.34 \\\hline\end{tabular} } ;
  
  \node[below=0.35cm of R4,xshift=10pt] (CPT4) {
    \begin{tabular}{|c|c|}\hline
      $G$ & $\prob(R_4|G)$ \\ \hline
      $\oplus$ & 0.20 \\\hline
      $\ominus$ & 0.75 \\\hline\end{tabular} } ;

  \node[right=0.5cm of R5,yshift=-10pt] (CPT5) {
    \begin{tabular}{|c|c|}\hline
      $G$ & $\prob(R_5|G)$ \\ \hline
      $\oplus$ & 0.95 \\\hline
      $\ominus$ & 0.03 \\\hline\end{tabular} } ;

\end{tikzpicture}
}
  \end{center}
  \caption{Running example.} \label{fig:ex01} 
\end{figure*}
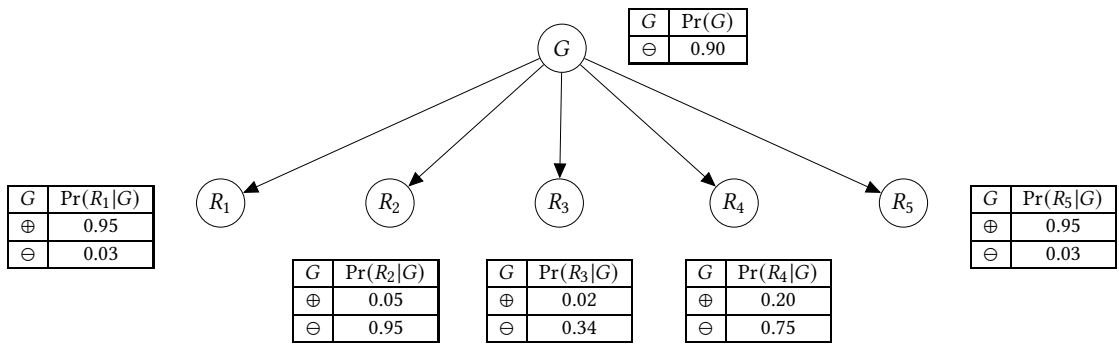

\paragraph{Running Example.}
Consider the NBC depicted graphically in \autoref{fig:ex01}~\footnote{%
      This example of an NBC is adapted from~\cite{msgcin-nips20}, which 
      is initially reported in~\cite[Ch.10]{barber-bk12}.}.
The features are the discrete random variables $R_1$, $R_2$, $R_3$,
$R_4$ and $R_5$. Each $R_i$ can take values $\lvt$ or $\lvf$ denoting,
respectively, whether a listener likes or not that radio station.
Random variable $G$ denotes an \tsf{age} class, which can take
values \tsf{Y} and \tsf{O}, denoting \tsf{young} and \tsf{older}
listeners, respectively. The target class $\oplus$ denotes the
prediction \emph{yes} (i.e.\ the listener likes the radio station) and
$\ominus$ denotes the prediction \emph{no} (i.e.\ the listerner does
not like the radio station). Thus, $\fml{K}=\{\ominus,\oplus\}$. 
Let us consider $\mbf{v}=(R_1,R_2,R_3,R_4,R_5)=(\lvt,\lvf,\lvf,\lvf,\lvt)$. 
We associate  $r_i$ to each literal ($R_i=\lvt$) and  $\neg{r_i}$ 
to literals ($R_i=\lvf$). 
Using~\eqref{eq:nbc5}, we get the values shown in~\autoref{fig:ex02}. 
(Note that to use positive values, we added $T=+4$ to each
$\lprob(\cdot)$.) 
As can be concluded, the classifier will predict $\oplus$.
\begin{figure*}[t]
  \begin{subfigure}[t]{\linewidth}
    \centering\scalebox{0.95}{
\renewcommand{\tabcolsep}{0.35em}
\renewcommand{\arraystretch}{1.175}
\begin{tabular}{|c|cccccc|c|} \hline  
  & $\prob(\oplus)$ & $\prob(r_1|\oplus)$ & $\prob(\neg{r_2}|\oplus)$ & 
  $\prob(\neg{r_3}|\oplus)$ & $\prob(\neg{r_4}|\oplus)$ & $\prob(r_5|\oplus)$ & 
  $\lprob(\oplus|\mbf{v})$
  \\ \hline  
  %
  $\prob(\cdot)$ & 0.10 & 0.95 & 0.95 & 0.98 & 0.80 & 0.95 &  
  \\
  %
  $\lprob(\cdot)$ & 1.70 & 3.95 & 3.95 & 3.98 & 3.78 & 3.95 & 21.31
  \\ \hline  
\end{tabular}
}
    \caption{Computing $\lprob(\oplus|\mbf{v})$}
  \end{subfigure}

  \bigskip
  \begin{subfigure}[t]{\linewidth}
    \centering\scalebox{0.95}{
\renewcommand{\tabcolsep}{0.35em}
\renewcommand{\arraystretch}{1.175}
\begin{tabular}{|c|cccccc|c|} \hline  
  & $\prob({\ominus})$ & $\prob(r_1|{\ominus})$ &
  $\prob(\neg{r_2}|{\ominus})$ & $\prob(\neg{r_3}|{\ominus})$ &
  $\prob(\neg{r_4}|{\ominus})$ & $\prob(r_5|{\ominus})$ &
  $\lprob(\ominus|\mbf{v})$
  \\ \hline  
  %
  $\prob(\cdot)$ & 0.90 & 0.03 & 0.05 & 0.66 & 0.25 &  0.03 &
  \\
  %
  $\lprob(\cdot)$ & 3.89 & 0.49 & 1.00 & 3.58 & 2.61 &  0.49 & 12.06
  \\ \hline  
\end{tabular}
}
    \caption{Computing $\lprob(\ominus|\mbf{v})$}
  \end{subfigure}
  \centering
  \caption{Deciding prediction for
    $\mbf{v}=(\lvt,\lvf,\lvf,\lvf,\lvt)$} \label{fig:ex02}
\end{figure*}
%

\section{Explaining NBCs in Polynomial Time}  \label{sec:xlc}
This section overviews the approach proposed in~\cite{msgcin-nips20}
for computing AXp's for binary NBCs. The general idea is to reduce the
NBC problem into an Extended Linear Classifier (XLC) and then explain
the resulting XLC. 
Our purpose is to devise a new approach that builds on XLC formulation 
to compute $\delta$-relevant sets for NBCs.  Hence, it is useful to recall 
first the translation of NBCs into XLCs and AXp's extraction from XLCs. 

\subsection{Extended Linear Classifiers}
We consider an XLC with categorical features. 
(Recall that the paper considers NBCs with binary classes and
categorical data.) 
Each feature $i \in \fml{F}$ has $x_i \in \{1, \dots, d_i\}$, (i.e.\ 
$\mbb{D}_i = \{1, \dots, d_i\}$).
Let,
\begin{equation} \label{eq:xlc01}
  \nu(\mbf{x})\triangleq%
  w_0 + 
  \sum\nolimits_{i\in\fml{F}}\sigma(x_i,v_i^1,v_i^2,\ldots,v_i^{d_i})
\end{equation}
$\sigma$ is a selector function that picks the value $v_i^{r}$ iff
$x_i$ takes value $r$.
Moreover, let us define the decision function, $\kappa(\mbf{x})=\oplus$
if $\nu(\mbf{x})>0$ and $\kappa(\mbf{x}) = \ominus $ if $\nu(\mbf{x})\le0$.

The reduction of a binary NBC, with categorical features, to an XLC 
is completed by setting:
$w_0\triangleq\lprob(\oplus)-\lprob(\ominus)$,
$v_i^1\triangleq\lprob(x_i=1|\oplus)-\lprob(x_i=1|\ominus)$,
$v_i^2\triangleq\lprob(x_i=2|\oplus)-\lprob(x_i=2|\ominus)$, 
$\dots$, $v_i^{d_i}\triangleq\lprob(x_i=d_i|\oplus)-\lprob(x_i=d_i|\ominus)$.
Hence, the argmax in~\eqref{eq:nbc5} is replaced with inequality to
get the following: 
%
%
\begin{align}\label{eq:nbc2xlc}
  \nonumber
  \lprob(\oplus) - \lprob(\ominus) +  \sum\nolimits_{i=1}^{m} \sum\nolimits_{k=1}^{k=d_j} & (\lprob(x_i=k|\oplus) ~ - \\ 
   & \lprob(x_i=k|\ominus))  (x_i = k) > 0 
\end{align}

\begin{figure*}[t]
  \begin{subfigure}[b]{0.5\linewidth}
    \centering
\renewcommand{\tabcolsep}{0.3em}
\renewcommand{\arraystretch}{1.225}
\begin{tabular}{|c|cc|cc|cc|cc|cc|} \hline  
  $w_0$ &
  $v_1^1$ & $v_1^2$ &
  $v_2^1$ & $v_2^2$ &
  $v_3^1$ & $v_3^2$ &
  $v_4^1$ & $v_4^2$ &
  $v_5^1$ & $v_5^2$
  \\ \hline  
  -2.19 &
  -2.97 & 3.46 &  
  2.95 & -2.95 &
  0.4 & -2.83 &
  1.17 & -1.32 &
  -2.97 & 3.46 
  \\ \hline  
\end{tabular}

    \caption{Example reduction of NBC to XLC (Example~\ref{ex:ex01})} \label{fig:nbc2xlc}
  \end{subfigure} 
  \begin{subfigure}[b]{0.49\linewidth}
    \centering\renewcommand{\tabcolsep}{0.3em}
\renewcommand{\arraystretch}{1.225}
%
\begin{tabular}{|c|ccccc|c|} \hline  
  $\Gamma$ &    
  $\delta_1$ &  
  $\delta_5$ & 
  $\delta_2$ &  
  $\delta_3$ &  
  $\delta_4$ &  
  $\Phi$        
  \\ \hline  
  9.25  &
  6.43 &
  6.43 & 
  5.90 &
  3.23 &
  2.49 &
  15.23
  \\ \hline  
\end{tabular}

    \caption{Computing $\delta_j$'s for the XLC 
      (Example~\ref{ex:ex02})} \label{fig:ex03}
  \end{subfigure}
  \caption{Values used in the running example (Example~\ref{ex:ex01}
    and Example~\ref{ex:ex02})} \label{fig:exs}
\end{figure*}

\begin{example} \label{ex:ex01}
  \autoref{fig:nbc2xlc} shows the resulting XLC formulation for the
  example in~\autoref{fig:ex02}. We also let $\lvf$ be associated
  with value 1 and $\lvt$ be associated with value 2, and $d_i=2$.
\end{example}

\subsection{Explaining XLCs}
We now describe how AXp's can be computed for XLCs. 
For a given instance $\mbf{x}=\mbf{a}$, define a \emph{constant} slack
(or gap) value $\Gamma$ given by,
\begin{equation} \label{eq:xlc02}
  \Gamma\triangleq\nu(\mbf{a})=%
                     \sum\nolimits_{i\in\fml{F}}\sigma(a_i,v_i^1,v_i^2,\ldots,v_i^{d_i})
\end{equation}
Computing an AXp corresponds to finding a 
subset-minimal set of literals $\fml{S}\subseteq\fml{F}$
such that~\eqref{eq:axp} holds, or alternatively,
\begin{equation} \label{eq:xlc03}
  \forall(\mbf{x}\in\mbb{F}).\bigwedge\nolimits_{i\in\fml{S}}(x_i=a_i) \ \limply \ \left(\nu(\mbf{x})>0\right)
\end{equation}
under the assumption that $\nu(\mbf{a})>0$. 
Thus, the purpose is to find  the \emph{smallest} slack
that can be achieved by allowing the feature not in $\fml{S}$ to
take any value (i.e. \emph{universal}/\emph{free} features), 
given that the literals in $\fml{S}$ are fixed by $\mbf{a}$ (i.e. 
$\bigwedge\nolimits_{i\in\fml{S}}(x_i=a_i)$).

Let $v_i^{\omega}$ denote the \emph{smallest} (or
\emph{worst-case}) value associated with $x_i$. 
Then, by letting every $x_i$ take \emph{any} value,
the \emph{worst-case} value of $\nu(\mbf{e})$ is,
\begin{equation}
  \Gamma^{\omega}=w_0+\sum\nolimits_{i\in\fml{F}}v_i^{\omega}
\end{equation}
Moreover, from~\eqref{eq:xlc02}, we have:  
$\Gamma=w_0+\sum_{i\in\fml{F}}v_i^{a_i}$.
The expression above can be rewritten
as follows,
\begin{equation}
  \begin{array}{rcl}
    \Gamma^{\omega} & = &
    w_0+\sum\nolimits_{i\in\fml{F}}v_i^{a_i}-\sum\nolimits_{i\in\fml{F}}(v_i^{a_i}-v_i^{\omega})\\[3.0pt]
    & = & \Gamma - \sum\nolimits_{i\in\fml{F}}\delta_i = -\Phi \\ 
  \end{array}
\end{equation}
where $\delta_i\triangleq{v_i^{a_i}}-{v_i^{\omega}}$,
and
$\Phi\triangleq\sum_{i\in\fml{F}}\delta_i-\Gamma=-\Gamma^{\omega}$.
Recall the goal is to find a  subset-minimal set $\fml{S}$ such
that the prediction is still $c$ 
(whatever the values of the other features):
\begin{equation} \label{eq:xlc04}
  w_0 + \sum\nolimits_{i \in \fml{S}} v_i^{a_i} + \sum\nolimits_{i \notin \fml{S}} v_i^{\omega} =
  -\Phi + \sum\nolimits_{i\in\fml{S}}\delta_i > 0
\end{equation}
In turn,~\eqref{eq:xlc04} can be represented as the following
knapsack problem~\cite{pisinger-bk04}:
\begin{equation} \label{eq:xlc05}
  \begin{array}{lcl}
    \tn{min}  & \quad & \sum_{i=1}^{m}p_i \\[4.5pt]
    \tn{such that} & \quad &
    \sum_{i=1}^{m}\delta_ip_i>\Phi \\[2.5pt]
    & & p_i\in\{0,1\}\\
  \end{array}
\end{equation}
where the variables $p_i$ assigned value 1 denote the indices included
in $\fml{S}$.
Note that, the fact that the coefficients
in the cost function are all equal to 1 makes the problem solvable in
log-linear time.
%
%

\begin{example} \label{ex:ex02}
  \autoref{fig:ex03} shows the values used for computing explanations
  for the example in~\autoref{fig:ex02}.
  For this example, the sorted $\delta_j$'s become
  $\langle\delta_1,\delta_5,\delta_2,\delta_4,\delta_3\rangle$.
  By picking $\delta_1$, $\delta_2$ and $\delta_5$, we ensure that 
  the prediction  is $\oplus$, independently of the values assigned 
  to features 3 and 4.
  Thus $\{1, 2, 5\}$ is an AXp for the NBC shown
  in~\autoref{fig:ex01}, with the input instance
  $(v_1,v_2,v_3,v_4,v_5)=(\lvt,\lvf,\lvf,\lvf,\lvt)$.
  (It is easy to observe that
  $\kappa((\lvt,\lvf,\lvf,\lvt,\lvt))=\kappa((\lvt,\lvf,\lvt,\lvf,\lvt))=\kappa((\lvt,\lvf,\lvt,\lvt,\lvt))=%
  \kappa((\lvt,\lvt,\lvf,\lvf,\lvt))=\kappa((\lvt,\lvt,\lvf,\lvt,\lvt))=\kappa((\lvt,\lvt,\lvt,\lvf,\lvt))=\oplus$.)
\end{example}

\section{$\delta$-Relevant Sets for NBCs}
\label{sec:paxp}

This section investigates the computation of $\delta$-relevant sets in
the concrete case of NBCs.

Observe that~\cref{def:mdrs} imposes no restriction on the
representation of the classifier that is assumed in earlier
work~\cite{kutyniok-jair21}, i.e.\ the logical representation of
$\kappa$ need not be a boolean circuit.
As a result, we extend  the definitions from earlier 
work~\cite{kutyniok-jair21}, as detailed below.

\subsection{Weak, Approximate \& Smallest Probabilistic AXp's}
A \emph{weak probabilistic $\axp$} ($\wdrset$) is a set of fixed 
features for which the conditional probability of predicting the
correct class $c$ exceeds $\delta$, given $c=\kappa(\mbf{v})$. 
Thus, $\fml{S}\subseteq\fml{F}$ is a  $\wdrset$ if,
%
\begin{align} \label{eq:wcdrs2}
  \wdrset&(\fml{S};\mbb{F},\kappa,\mbf{v},\delta)  
  \nonumber \\
  :=\,\: & \prob_{\mbf{x}}(\kappa(\mbf{x})=c\,|\,\mbf{x}_{\fml{S}}=\mbf{v}_{\fml{S}})
  \ge \delta
  \\[1.0pt]
  :=\,\: &\frac{%
    |\{\mbf{x}\in\mbb{F}:\kappa(\mbf{x})=c\land(\mbf{x}_{\fml{S}}=\mbf{v}_{\fml{S}})\}|
  }{%
    |\{\mbf{x}\in\mbb{F}:(\mbf{x}_{\fml{S}}=\mbf{v}_{\fml{S}})\}|
  }
  \ge\delta \nonumber
\end{align}
which means that the fraction of the number of models predicting the
target class and consistent with the fixed features (represented by
$\fml{S}$), given the total number of points in feature space
consistent with the fixed features, must exceed $\delta$.
(The main difference to~\eqref{eq:drs} is that features and classes
are no longer required to be boolean. Also, the definition makes
explicit the parameterizations assumed.)
Moreover, a \emph{probabilistic $\axp$} ($\drset$) $\fml{X}$ is a
$\wdrset$ that is also subset-minimal, 
\begin{align}
  \drset&(\fml{X};\mbb{F},\kappa,\mbf{v},\delta) \::= \nonumber \\
  &\wdrset(\fml{X};\mbb{F},\kappa,\mbf{v},\delta) \:\:\land \\
  &\forall(\fml{X}'\subsetneq\fml{X}). %
  \neg\wdrset(\fml{X}';\mbb{F},\kappa,\mbf{v},\delta) \nonumber
\end{align}
Minimum-size $\drset$'s ($\mdrset$, or smallest $\drset$) generalize
Min-$\delta$-relevant sets in~\cref{def:mdrs}.
Furthermore, we define an \emph{approximate} 
probabilistic AXp  ($\adrset$)  $\fml{X}$ as a $\wdrset$ such that 
the removal of any single feature $i$ from $ \fml{X}$ will falsify   
$ \wdrset(\fml{X}\setminus\{i\};\mbb{F},\kappa,\mbf{v},\delta)$. 
Formally:
\begin{align}
  \adrset&(\fml{X};\mbb{F},\kappa,\mbf{v},\delta) \::= \nonumber \\
  &\wdrset(\fml{X};\mbb{F},\kappa,\mbf{v},\delta) \:\:\land \\
  &\forall(\fml{X}\setminus\{i\}). %
  \neg\wdrset(\fml{X}\setminus\{i\};\mbb{F},\kappa,\mbf{v},\delta) \nonumber
\end{align}

As stated earlier, the main purpose of this paper is to investigate
the computation of $\adrset$ explanations. The next section introduces
a pseudo-polynomial time algorithm for computing  $\adrset$'s.
Although, $\adrset$ are not minimal subset/cardinality,  our
experiments show that the proposed approach computes (in
pseudo-polynomial time) succinct~\cite{miller-pr56} and highly
precise approximate explanations.

\subsection{Counting Models of XLCs} 

Earlier
work~\cite{dyer-stoc03,klivans-focs11,weimann-icalp18,tomescu-ic19}
proposed the use of dynamic programming (DP) for approximating the
number of feasible solutions of the 0-1 knapsack constraint, i.e.\ the
\#knapsack problem.
Here we propose an extension of the basic formulation, to allow
counting feasible solutions of XLCs.

%
We are interested in the number of solutions of,
\begin{equation} \label{eq:dp01}
  \sum\nolimits_{j\in\fml{F}}\sigma(x_j,v_j^1,v_j^2,\ldots,v_j^{d_j})>-w_0
\end{equation}
where we assume all $v^i_j$ to be integer-valued, and non-negative
(e.g.\ this is what the translation from NBCs to XLCs yields).
Moreover,~\eqref{eq:dp01} can be written as follows:

\begin{equation} \label{eq:dp02}
  \sum\nolimits_{j\in\fml{F}}\sigma(x_j,-v_j^1,-v_j^2,\ldots,-v_j^{d_j})<{w_0}
\end{equation}
which reveals the relationship with the Knapsack constraint.

For each $j$, let us sort the $-v_j^i$ in non-decreasing order,
collapsing duplicates, and counting the number of duplicates,
obtaining two sequences:
\[
\begin{array}{l}
  \langle w^1_j,\ldots,w^{d^{'}_j}_j\rangle \\[4.5pt]
  \langle n^1_j,\ldots,n^{d^{'}_j}_j\rangle \\
\end{array}
\]
such that $w^1_j<w^2_j<\ldots<w^{d^{'}_j}_j$ and each $n^i_j\ge1$
gives the number of repetitions of weight $w^i_j$.


\paragraph{Counting.}
Let $C(k,r)$ denote the number of solutions of~\eqref{eq:dp02} when
the subset of features considered is $\{1,\ldots,k\}$ and the sum of
picked weights is at most $r$.
To define the solution for the first $k$ features, taking into account
the solution for the first $k-1$ features, we must consider that the
solution for $r$ can be obtained due to \emph{any} of the possible
values of $x_j$.
As a result, for an XLC 
the general recursive definition of $C(k,r)$ becomes,
\[
C(k,r)= \sum\nolimits_{i=1}^{d^{'}_k}n^i_{k}\times{C}(k-1,r-w^i_{k})
\]

Moreover, $C(1,r)$ is given by,
\[
C(1,r)=\left\{%
\begin{array}{lcl}
  0 & \quad & \tn{if $r<w^1_1$} \\[4.5pt]
  n^1_1 & \quad & \tn{if $w^1_1\le r<w^2_1$} \\[5.5pt]
  n^1_1+n^2_1 & \quad & \tn{if $w^2_1\le r<w^3_1$} \\[2pt]
  \ldots \\[2pt]
  \sum\nolimits_{i=1}^{d^{'}_1}n^i_1 & \quad & \tn{if $w^{d^{'}_1}_1\le r$} \\[2pt]
\end{array}
\right.
\]
In addition, if $r<0$, then $C(k,r)=0$, for $k=1,\ldots,m$.
Finally, the dimensions of the $C(k,r)$ table are as follows:
\begin{enumerate}
\item The number of rows is $m$.
\item The (worst-case) number of columns is given by:
  \begin{equation} \label{eq:wval}
    W'=\sum_{j\in\fml{F}}{n^{d'_j}_j}\times{w^{d'_j}_j}
  \end{equation}
  $W'$ represents the largest possible value, in theory. However, in
  practice, it suffices to set the number of columns to $W=w_0+T$,
  which is often much smaller than $W'$.
\end{enumerate}


\begin{example} \label{ex:05}
  Consider the following problem. There are 4 features,
  $\fml{F}=\{1,2,3,4\}$. Each feature $j$ takes values in
  $\{1,2,3\}$, i.e.\ $x_j\in\{1,2,3\}$. The prediction should be 1
  when the sum of the values of the $x_j$ variables is no less than 8.
  We set $w_0 = -7$, and get 
  the formulation,
  \[
  \sum_{j\in\{1,2,3,4\}}\sigma(x_j,1,2,3)>7
  \]
  where each $x_j$ picks value in $\{ 1,2,3 \}$.
  We translate to the extended knapsack formulation and obtain:
  \[
  \sum_{j\in\{1,2,3,4\}}\sigma(x_j,-1,-2,-3)<-7
  \]
  We require the weights to be integer and non-negative, and so we sum
  to each $w_j^k$ the complement of the most negative $w_j^k$ plus 1. 
  Therefore, we add +4 to each $j$ and +16 to right-hand side of the inequality. 
  Thus, we get
  \[
  \sum_{j\in\{1,2,3,4\}}\sigma(x_j,3,2,1)<9
  \]
  For this formulation, $x_j=1$ picks value $3$. (For example, we can
  pick two $x_j$ with value 1, but not 3, as expected.)
  
  In this case, the DP table size will be $4\times12$, even though we
  are interested in entry $C(4,8)$.
  \begin{table}[t]
  \begin{center}
    \renewcommand{\arraystretch}{1.125}
    \renewcommand{\tabcolsep}{0.5em}
    \begin{tabular}{|c||c|c|c|c|c|c|c|c|c|c|c|c|c|}\hline
      \diagbox{$k$}{$r$} 
      & 0 & 1 & 2 & 3 & 4 & 5 & 6 & 7 & 8 & 9 & 10 & 11 & 12
      \\ \hline\hline
      1 & 0 & 1 & 2 & 3 & 3 & 3 & 3 & 3 & 3 & -- & -- & -- & --
      \\ \hline
      2 & 0 & 0 & 1 & 3 & 6 & 8 & 9 & 9 & 9 & -- & -- & -- & --
      \\ \hline
      3 & 0 & 0 & 0 & 1 & 4 & 10 & 17 & 23 & 16 & -- & -- & -- & --
      \\ \hline
      4 & 0 & 0 & 0 & 0 & 1 & 5 & 15 & 31 & 50 & -- & -- & -- & --
      \\ \hline
    \end{tabular}
    \caption{DP table for Example~\ref{ex:05}} \label{tab:ex05}
  \end{center}
\end{table}

  \autoref{tab:ex05} shows DP table, and the number of solutions for
  the starting problem, i.e.\ there are 50 combinations of values
  whose sum is no less than 8.
\end{example}

By default, the dynamic programming formulation assumes that features
can take any value. However, the same formulation can be adapted when
features take a given (fixed) value. Observe that this will be
instrumental for computing $\adrset$'s.

Consider that feature $k$ is fixed to value $l$. Then, the formulation
for $C(k,r)$ becomes:
\[
C(k,r)=n^l_k\times{C}(k-1,r-w^l_k)={C}(k-1,r-w^l_k)
\]
Given that $k$ is fixed, then it is the case that $n^l_k=1$.
%

\begin{example} \label{ex:05b}
  For Example~\ref{ex:05}, assume that $x_2=1$ and $x_4=3$.
  Then, the constraint we want to satisfy is:
  \[
  \sum_{j\in\{1,3\}}\sigma(x_j,1,2,3)>3
  \]
  Following a similar transformation into knapsack formulation, we get 
  \[ \sum_{j\in\{1,3\}}\sigma(x_j,3,2,1) < 5 \]
  
  After updating the DP table, with fixing features 2 and 4, we get 
  the DP table shown in~\autoref{tab:ex05b}. As a result,  
  we can conclude that the number of solutions is 6.
\end{example}
\begin{table}[t]
  \begin{center}
    \renewcommand{\arraystretch}{1.125}
    \renewcommand{\tabcolsep}{0.5em}
    \begin{tabular}{|c||c|c|c|c|c|c|c|c|c|c|c|c|c|}\hline
      \diagbox{$k$}{$r$} 
      & 0 & 1 & 2 & 3 & 4 & 5 & 6 & 7 & 8 & 9 & 10 & 11 & 12
      \\ \hline\hline
      1 & 0 & 1 & 2 & 3 & 3 & 3 & 3 & 3 & 3 & -- & -- & -- & --
      \\ \hline
      2 & 0 & 0 & 0 & 0 & 1 & 2 & 3 & 3 & 3 & -- & -- & -- & --
      \\ \hline
      3 & 0 & 0 & 0 & 0 & 0 & 1 & 3 & 6 & 8 & -- & -- & -- & --
      \\ \hline
      4 & 0 & 0 & 0 & 0 & 0 & 0 & 1 & 3 & 6 & -- & -- & -- & --
      \\ \hline
    \end{tabular}
    \caption{DP table for Example~\ref{ex:05b}} \label{tab:ex05b}
  \end{center}
\end{table}

The table $C(k,r)$ can be filled out in pseudo-polynomial time.
The number of rows is $m$. The number of columns is
$W$ (see~\eqref{eq:wval}). Moreover, the computation of each entry
uses the values of at most $m$ other entries.
Thus, the total running time is:
$\Theta(m^2\times{W})$.

\paragraph{From XLCs to Positive Integer Knapsacks.}
To assess heuristic explainers, we consider NBCs, and use a
standard transformation from probabilities to positive real
values~\cite{park-aaai02}.
Afterwards, we convert the real values to integer values by scaling
the numbers. However, to avoid building a very large DP table, we
implement the following optimization.
The number of decimal places of the probabilities is reduced while
there is no decrease in the accuracy of the classifier both on
training and on test data. In our experiments, we observed that there
is no loss of accuracy if four decimal places are used, and that there
is a negligible loss of accuracy with three decimal places.

\paragraph{Assessing  explanation precision.}
Given a Naive Bayes classifier, expressed as an XLC, we can assess 
explanation accuracy in pseudo-polynomial time.
Given an instance $\mbf{v}$, a prediction $\kappa(\mbf{v})=\oplus$, and 
an approximate explanation $\mbf{S}$, we can use the approach described 
in this section to count the number of 
instances consistent with the explanation for which the prediction
remains unchanged (i.e.\  number of points $\mbf{x}\in\mbb{F}$ s.t. 
$(\kappa(\mbf{x})=\kappa(\mbf{v})\land(\mbf{x}_{\fml{S}}=\mbf{v}_{\fml{S}}))$). 
Let this number be $n_{\oplus}$ (given the
assumption that the prediction is $\oplus$). Let the number of
instances with a different prediction ($\ominus\neq\kappa(\mbf{v})$)\footnote{%
As we are in binary setting, then $\ominus=\neg\oplus=\neg\kappa(\mbf{v})$).} 
be  $n_{\ominus}$. 
Hence, the conditional probability~\eqref{eq:pdefs} can be defined, in 
the case of NBCs, as follow:

\[ %
	\prob_{\mbf{x}}(\kappa(\mbf{x})=\oplus\,|\,\mbf{x}_{\fml{S}}=\mbf{v}_{\fml{S}})  = 
 	\frac{ n_{\oplus} }{  |\{\mbf{x}\in\mbb{F}:(\mbf{x}_{\fml{S}}=\mbf{v}_{\fml{S}})\}| } %
\]

Observe that the numerator %
$ |\{\mbf{x}\in\mbb{F}:\kappa(\mbf{x})=\oplus\land(\mbf{x}_{\fml{S}}=\mbf{v}_{\fml{S}})\}|$ is 
expressed  by  the number of models $n_\oplus$, i.e.\ the points
$\mbf{x}$ in feature space that are consistent with $\mbf{v}$ given
$\fml{S}$ and with prediction $\oplus$. Further, we have 

\begin{align*}
\prob_{\mbf{x}}(\kappa(\mbf{x})=\oplus\,|\,\mbf{x}_{\fml{S}}=\mbf{v}_{\fml{S}})  = &
 1 - \prob_{\mbf{x}}(\kappa(\mbf{x})=\ominus\,|\,\mbf{x}_{\fml{S}}=\mbf{v}_{\fml{S}})  \\
 = & 1 - \frac{ n_{\ominus} }{  |\{\mbf{x}\in\mbb{F}:(\mbf{x}_{\fml{S}}=\mbf{v}_{\fml{S}})\}| } 
\end{align*} 

%

where $n_\ominus = |\{\mbf{x}\in\mbb{F}:\kappa(\mbf{x})=\ominus\land(\mbf{x}_{\fml{S}}=\mbf{v}_{\fml{S}})\}|$.

\subsection{Computing ApproxPAXp's}
\cref{alg:apaxp} depicts our method for computing $\adrset$'s 
given a prediction function $\kappa$ of an NBC, an input instance 
$\mbf{v}$ and a threshold $\delta$. 
The procedure ApproxPAXp is referred to as a deletion-based algorithm%
\footnote{This sort of algorithm can be traced at least to the work of
  Valiant\cite{valiant-cacm84}, but some authors~\cite{juba-aaai16}
  argue that it is also implicit in works from the
  $\text{19}^{\text{th}}$ century~\cite{mill-bk43}.}; it starts from a
set of features $\fml{S}$, e.g.\ initialized to $\fml{F}$ and then it
iteratively drops features while the updated set $\fml{S}$ remains a
$\wdrset$. 
The function $\isweakpaxp$ implements the approach described in the
previous section, which measures explanation precision by exploiting a
pseudo-polynomial algorithm for model counting. Hence, it is implicit
that the DP table is updated at each iteration of the loop in the
ApproxPAXp procedure.
More specifically, when a feature $i$ is newly set universal, its associated 
cells  $C(i,r)$ are recalculated such that 
$C(k,r)= \sum\nolimits_{i=1}^{d^{'}_k}n^i_{k}\times{C}(k-1,r-w^i_{k})$;  
and when $i$ is fixed, i.e.\ $i \in \fml{S}$, then $C(i,r) = C(i-1, r-v^j_i)$ where  
$v^j_i\triangleq\lprob(v_i=j|c)-\lprob(v_i=j|\neg c)$.  
Furthermore, we point out that in our experiment,  $\fml{S}$ is initialized 
to an AXp $\fml{X}$  that we compute initially for all tested 
instances using the outlined (polynomial) algorithm in~\cref{sec:xlc}.
It is easy to observe that features not belonging to $\fml{X}$ do not contribute 
in the decision of $\kappa(\mbf{v})$ (i.e.\ their removal does not change 
the value of $n_\ominus$ that equals to zero) and thus can be set universal 
at the initialisation step, which allows us to improve the performance of 
\cref{alg:apaxp}.   

Moreover, we apply an heuristic order over $\fml{S}$ that
aims to remove earlier less relevant features and thus to produce  
shorter  approximate explanations.  Typically, we order $\fml{S}$ 
following the increasing order of $\delta_i$ values, namely the reverse 
order applied to compute the AXp.    
Conducted preliminary experiments using a (naive heuristic)
lexicographic order over the features show less succinct explanations. 

Finally, notice that~\cref{alg:apaxp} can be used to compute 
an AXp, i.e.\ an $\adrset$ with $\delta = 1$. Nevertheless, the polynomial 
time algorithm for computing AXp's proposed in~\cite{msgcin-nips20} 
remains a better choice to use in case of AXp's than \cref{alg:apaxp} which 
runs in pseudo  polynomial time.

\begin{algorithm}[t]
  \begin{flushleft}
\hspace*{\algorithmicindent}
\textbf{Input}: {Classifier $\kappa$, instance $\mbf{v}$, threshold $\delta$}\\
\hspace*{\algorithmicindent}
\textbf{Output}: { $\adrset~~\fml{S}$}
\end{flushleft}
\begin{algorithmic}[1]
  \Procedure{$\adrset$}{$\kappa,\mbf{v}, \delta$}
  \State{$\fml{S} \gets \{1,\ldots,m\}$}
  \For{$i\in\{1,\ldots,m\}$}
    \If{$\isweakpaxp(\fml{S}\setminus\{i\}, \kappa(\mbf{x})=c, \delta)$}
      \State{$\fml{S} \gets \fml{S}\setminus\{i\}$}
    \EndIf
  \EndFor  
  \State{\bfseries{return}~{$\fml{S}$}}
\EndProcedure
\end{algorithmic}
  \caption{Computing an $\adrset$ 
  }
  \label{alg:apaxp}
\end{algorithm}

\begin{example} \label{ex:06}
Let us consider again the NBC of the running example (Example~\ref{ex:ex01}) and 
$\mbf{v}=(\lvt,\lvf,\lvf,\lvf,\lvt)$. The corresponding XLC 
is shown in~\autoref{fig:ex03} (Example~\ref{ex:ex02}). Also, consider 
the AXp $\{1,  2, 5\}$ of $\mbf{v}$ and $\delta = 0.85$. 
The resulting DP table for $\fml{S} = \{1,  2, 5\}$ is shown 
in~\autoref{tab:ex06a}. Note that for illustrating small tables, we set 
the number of decimal places to zero (greater number of decimal 
places, i.e.\ 1,2, etc, were tested and return the results).
(Also, note that the DP table reports  ``\textemdash'' if the cell is not 
calculated during the running of \cref{alg:apaxp}.)    
Moreover, we convert  the probabilities into positive integers, 
so we sum to each $w_j^k$ the complement of the most 
negative $w_j^k$ plus 1. 
The resulting weights are shown in~\autoref{fig:ex06}. 
Thus, we get 
$\sum\nolimits_{i\in\{1,2,3,4,5\}} \sigma(x_i, w_i^1, w_i^2) < 17$. 
%
Observe that the number of models $n_\oplus = C(5,16)$, and 
$C(5,16)$ is calculated using $C(4,16-w^2_5)=C(4,15)$, i.e.\ $C(4,15)=C(5,16)$ 
(feature 5 is fixed, so it is allowed to take only the value $w^2_5=1$). Next, 
$C(4, 15)  = C(3, 15-w^1_4 )  + C(3, 15-w^2_4) = C(3, 12)  + C(3, 14)$ 
(feature 4 is free, so it is allowed to take any value of $\{w^1_4,w^2_4\}$); 
the recursion ends when k=1, namely for $C(1,5) = C(2,6) = n^2_1 = 1$, 
$C(1,7) = C(2,7) = n^2_1 = 1$, $C(1,8) = C(2,8) = n^2_1 = 1$ and 
$C(1,10) = C(2,11) = n^2_1 = 1$ (feature 1 is fixed and takes value $w^2_1$).
Next, \autoref{tab:ex06b} (resp.\  \autoref{tab:ex06c} and \autoref{tab:ex06d}) report 
the resulting DP table for $\fml{S} = \{2,5\}$ (resp.\  $\fml{S} = \{1,5\}$ and 
$\fml{S} = \{1\}$). 
It is easy to confirm that after dropping feature 2, the precision 
of $\fml{S}= \{1,5\}$ becomes $87.5\%$, i.e.\  $\frac{7}{8} = 0.875 > \delta$.
Furthermore, observe that the resulting  $\fml{S}$ when dropping feature 1 or 
2 and 5, are not $\wdrset$'s, namely,  the precision of $\{2,5\}$  is 
$\frac{6}{8} = 0.75 < \delta$ and the precision of $\{1\}$ is 
$\frac{9}{16} = 0.5625 < \delta$. 
In summary, \cref{alg:apaxp} starts with $\fml{S} = \{1, 2,5\}$, then at  
iteration\#1, feature 1 is tested and since  $\{2,5\}$ is not $\wdrset$ then 
1 is kept in $\fml{S}$; at iteration\#2, feature 2 is tested and 
since $\{1,5\}$  is a $\wdrset$, then $\fml{S}$ is updated (i.e.\ $\fml{S}= \{1,5\}$); 
at iteration\#3, feature 5 is tested and since  $\{1\}$ is not  a $\wdrset$, then
 5 is saved in $\fml{S}$.
As a result, the delivered $\adrset$ is $\{1,5\}$.
\end{example}

Let us underline that we could initialize $\fml{S}$ to $\fml{F}$, in which 
case the number of models would be 1. However, we opt instead 
to always start from an AXp. 
In the example, the AXp is $\{1,2,5\}$  which, because it is an AXp, 
the number of models must be 4 (i.e. $2^2$, since two features are free).

For any proper subset of the AXp, with $r$ free variables, it must be the case 
that the number of models is strictly less than $2^r$. Otherwise, 
we would have an AXp as a proper subset of another AXp; 
but this would contradict the definition of AXp. 
The fact that the number of models is strictly less than $2^r$ is confirmed 
by the examples of subsets considered. 
It must also be the case that if  $\fml{S}'\subseteq\fml{S}$, then the number 
of models of $\fml{S}'$ must not exceed the number of models of  $\fml{S}$. 
So, we can argue that there is monotonicity in the number of models, but not 
on the precision.

\begin{figure}[b]
  \centering
  
\renewcommand{\tabcolsep}{0.3em}
\renewcommand{\arraystretch}{1.225}
\begin{tabular}{|c|cc|cc|cc|cc|cc|} \hline  
  $W$ &
  $w_1^1$ & $w_1^2$ &
  $w_2^1$ & $w_2^2$ &
  $w_3^1$ & $w_3^2$ &
  $w_4^1$ & $w_4^2$ &
  $w_5^1$ & $w_5^2$
  \\ \hline  
  16 &
  7 & 1 &  
  1 & 6 &
  3 & 6 &
  1 & 3 &
  7 & 1 
  \\ \hline  
\end{tabular}

  \caption{\#knapsack problem of Example~\ref{ex:06}} \label{fig:ex06} 
\end{figure}
%

\begin{table*}[tb]
  \begin{center}
    \renewcommand{\arraystretch}{1.125}
    \renewcommand{\tabcolsep}{0.5em}
    \begin{tabular}{|c||c|c|c|c|c|c|c|c|c|c|c|c|c|c|c|c|c|}\hline
      \diagbox{$k$}{$r$} 
      & 0 & 1 & 2 & 3 & 4 & 5 & 6 & 7 & 8 & 9 & 10 & 11 & 12 & 13 & 14 & 15 & 16
      \\ \hline\hline
1 & 0 & \textemdash & \textemdash & \textemdash & \textemdash & 1 & \textemdash & 1 & 1 & \textemdash & 1 	& \textemdash & \textemdash & \textemdash & \textemdash & \textemdash & \textemdash\\
	\hline
2 & 0 & \textemdash & \textemdash & \textemdash & \textemdash & \textemdash & 1 & \textemdash & 1 & 1 & 	\textemdash & 1 & \textemdash & \textemdash & \textemdash & \textemdash & \textemdash\\
	\hline
3 & 0 & \textemdash & \textemdash & \textemdash & \textemdash & \textemdash & \textemdash & \textemdash 	& \textemdash & \textemdash & \textemdash & \textemdash & 2 & \textemdash & 2 & \textemdash & \textemdash\\
	\hline
4 & 0 & \textemdash & \textemdash & \textemdash & \textemdash & \textemdash & \textemdash & \textemdash 	& \textemdash & \textemdash & \textemdash & \textemdash & \textemdash & \textemdash & \textemdash & 4 & 		\textemdash\\
	\hline
5 & 0 & \textemdash & \textemdash & \textemdash & \textemdash & \textemdash & \textemdash & \textemdash 	& \textemdash & \textemdash & \textemdash & \textemdash & \textemdash & \textemdash & \textemdash & \textemdash & 4\\
	\hline  
    \end{tabular}
    \caption{DP table for $\fml{S} = \{1, 2, 5\}$ (Example~\ref{ex:06})} \label{tab:ex06a}
  \end{center}
\end{table*}

\begin{table*}[tb]
  \begin{center}
    \renewcommand{\arraystretch}{1.125}
    \renewcommand{\tabcolsep}{0.5em}
    \begin{tabular}{|c||c|c|c|c|c|c|c|c|c|c|c|c|c|c|c|c|c|}\hline
      \diagbox{$k$}{$r$} 
      & 0 & 1 & 2 & 3 & 4 & 5 & 6 & 7 & 8 & 9 & 10 & 11 & 12 & 13 & 14 & 15 & 16
      \\ \hline\hline
1 & 0 & \textemdash & \textemdash & \textemdash & \textemdash & 1 & \textemdash & 1 & 2 & \textemdash & 2 & \textemdash & \textemdash & \textemdash & \textemdash & \textemdash & \textemdash\\
\hline
2 & 0 & \textemdash & \textemdash & \textemdash & \textemdash & \textemdash & 1 & \textemdash & 1 & 2 & \textemdash & 2 & \textemdash & \textemdash & \textemdash & \textemdash & \textemdash\\
\hline
3 & 0 & \textemdash & \textemdash & \textemdash & \textemdash & \textemdash & \textemdash & \textemdash & \textemdash & \textemdash & \textemdash & \textemdash & 3 & \textemdash & 3 & \textemdash & \textemdash\\
\hline
4 & 0 & \textemdash & \textemdash & \textemdash & \textemdash & \textemdash & \textemdash & \textemdash & \textemdash & \textemdash & \textemdash & \textemdash & \textemdash & \textemdash & \textemdash & 6 & \textemdash\\
\hline
5 & 0 & \textemdash & \textemdash & \textemdash & \textemdash & \textemdash & \textemdash & \textemdash & \textemdash & \textemdash & \textemdash & \textemdash & \textemdash & \textemdash & \textemdash & \textemdash & 6\\
\hline
    \end{tabular}
    \caption{DP table for $\fml{S} = \{2, 5\}$ (Example~\ref{ex:06})} \label{tab:ex06b}
  \end{center}
\end{table*}

\begin{table*}[tb]
  \begin{center}
    \renewcommand{\arraystretch}{1.125}
    \renewcommand{\tabcolsep}{0.5em}
    \begin{tabular}{|c||c|c|c|c|c|c|c|c|c|c|c|c|c|c|c|c|c|}\hline
      \diagbox{$k$}{$r$} 
      & 0 & 1 & 2 & 3 & 4 & 5 & 6 & 7 & 8 & 9 & 10 & 11 & 12 & 13 & 14 & 15 & 16
      \\ \hline\hline
1 & 0 & \textemdash & 1 & 1 & \textemdash & 1 & \textemdash & 1 & 1 & \textemdash & 1 & \textemdash & \textemdash & \textemdash & \textemdash & \textemdash & \textemdash\\
\hline
2 & 0 & \textemdash & \textemdash & \textemdash & \textemdash & \textemdash & 1 & \textemdash & 2 & 2 & \textemdash & 2 & \textemdash & \textemdash & \textemdash & \textemdash & \textemdash\\
\hline
3 & 0 & \textemdash & \textemdash & \textemdash & \textemdash & \textemdash & \textemdash & \textemdash & \textemdash & \textemdash & \textemdash & \textemdash & 3 & \textemdash & 4 & \textemdash & \textemdash\\
\hline
4 & 0 & \textemdash & \textemdash & \textemdash & \textemdash & \textemdash & \textemdash & \textemdash & \textemdash & \textemdash & \textemdash & \textemdash & \textemdash & \textemdash & \textemdash & 7 & \textemdash\\
\hline
5 & 0 & \textemdash & \textemdash & \textemdash & \textemdash & \textemdash & \textemdash & \textemdash & \textemdash & \textemdash & \textemdash & \textemdash & \textemdash & \textemdash & \textemdash & \textemdash & 7\\
\hline    
    \end{tabular}
    \caption{DP table for $\fml{S} = \{1, 5\}$ (Example~\ref{ex:06})} \label{tab:ex06c}
  \end{center}
\end{table*}

\begin{table*}[tb]
  \begin{center}
    \renewcommand{\arraystretch}{1.125}
    \renewcommand{\tabcolsep}{0.5em}
    \begin{tabular}{|c||c|c|c|c|c|c|c|c|c|c|c|c|c|c|c|c|c|}\hline
      \diagbox{$k$}{$r$} 
      & 0 & 1 & 2 & 3 & 4 & 5 & 6 & 7 & 8 & 9 & 10 & 11 & 12 & 13 & 14 & 15 & 16
      \\ \hline\hline
1 & 0 & 0 & 1 & 1 & 1 & 1 & \textemdash & 1 & 1 & \textemdash & 1 & \textemdash & \textemdash & \textemdash & \textemdash & \textemdash & \textemdash\\
\hline
2 & 0 & \textemdash & 0 & 1 & \textemdash & 1 & 1 & \textemdash & 2 & 2 & \textemdash & 2 & \textemdash & \textemdash & \textemdash & \textemdash & \textemdash\\
\hline
3 & 0 & \textemdash & \textemdash & \textemdash & \textemdash & \textemdash & 1 & \textemdash & 1 & \textemdash & \textemdash & \textemdash & 3 & \textemdash & 4 & \textemdash & \textemdash\\
\hline
4 & 0 & \textemdash & \textemdash & \textemdash & \textemdash & \textemdash & \textemdash & \textemdash & \textemdash & 2 & \textemdash & \textemdash & \textemdash & \textemdash & \textemdash & 7 & \textemdash\\
\hline
5 & 0 & \textemdash & \textemdash & \textemdash & \textemdash & \textemdash & \textemdash & \textemdash & \textemdash & \textemdash & \textemdash & \textemdash & \textemdash & \textemdash & \textemdash & \textemdash & 9\\
\hline
    \end{tabular}
    \caption{DP table for $\fml{S} = \{1\}$ (Example~\ref{ex:06})} \label{tab:ex06d}
  \end{center}
\end{table*}

\paragraph{Properties of $\adrset$'s.}
In addition to the comments above, and by carefully computing
$\adrset$'s, these can exhibit important properties.
Let $\fml{X}$ denote an AXp. Then, for any $\adrset$ $\fml{A}$
obtained using $\fml{X}$ as the seed, i.e.\ $\fml{A}$ is required to
be a subset of $\fml{X}$, then we have the following properties:
\begin{enumerate}[nosep]
\item $\fml{A}\subseteq\fml{X}$;
\item There exists a probabilistic abductive explanation $\fml{E}$
  such that $\fml{E}\subseteq\fml{A}$; and
\item $\fml{A}$ is a $\delta$-relevant set (see~\cref{def:drs}).
\end{enumerate}
Thus, an $\adrset$ $\fml{A}$ can be made to be a superset of some
$\drset$, a subset of some AXp, and such that $\fml{A}$ exhibits the
strong probabilistic properties of $\delta$-relevant sets.

\jnoteF{Properties of $\adrset$'s.}

\subsection{Related Work}
$\delta$-relevant set~\cite{kutyniok-jair21} were originally proposed
for computing probabilistic explanations on binary classifiers.
A concept related with $\delta$-relevant sets is that of Probabilistic
Sufficient Explanation~\cite{vandenbroeck-ijcai21}.
In contrast to $\delta$-relevant sets, which offer provably succinct
and precise probabilistic explanations under a uniform distribution
assumption, \cite{vandenbroeck-ijcai21} proposes an heuristic approach
for computing sufficient explanations, and leverage on measuring the
probability guarantee of getting the target class,  under a data
distribution. 
A recent alternative is the work of~\cite{tan-nips21}, which proposes
to compute
provably succinct and precise explanations, referred as 
\emph{size-$k$ $\epsilon$-error}\footnote{\emph{size-$k$
    $\epsilon$-error} explanations can be viewed as top-$k$
  $\delta$-relevant sets (i.e.\ parameterized with a size $k$).},  
for explaining surrogate decision tree classifiers that approximate
black-box models.

\setlength{\tabcolsep}{4pt}
\let\lpr\undefined
\let\rpr\undefined
\newcommand{\lpr}{(}
\newcommand{\rpr}{)}

\begin{table*}[t]
\centering
\resizebox{\textwidth}{!}{
  \begin{tabular}{l S[table-format=2]S[table-format=3]
  			S[table-format=2.2]  c S[table-format= 2]  c c S[table-format=3] S[table-format=1.3]  
			c c S[table-format= 3] S[table-format=1.3]   c c S[table-format= 3] S[table-format= 1.3]}
\toprule[1.2pt]
\multirow{2}{*}{\bf Dataset} & \multicolumn{2}{c}{\multirow{2}{*}{\bf (\#F \, \#I)}}  &  {\bf NBC}  &  {\bf AXp}  &  
& \multicolumn{4}{c}{\bf ApproxPAXp$_{\le 9}$}  & \multicolumn{4}{c}{\bf ApproxPAXp$_{\le 7}$}  & \multicolumn{4}{c}{\bf ApproxPAXp$_{\le 4}$} \\
\cmidrule[0.8pt](lr{.75em}){4-4}
\cmidrule[0.8pt](lr{.75em}){5-5}
\cmidrule[0.8pt](lr{.75em}){7-10}
\cmidrule[0.8pt](lr{.75em}){11-14}
\cmidrule[0.8pt](lr{.75em}){15-18}
& \multicolumn{2}{c}{}  & {\bf A\%} &  {\bf Length}  & {$\delta$}  &  {\bf Length} & {\bf Precision} & {\bf W\%} & {\bf Time}  &  
{\bf Length} & {\bf Precision} & {\bf W\%} &  {\bf Time} &  {\bf Length} & {\bf Precision} & {\bf W\%} &  {\bf Time}  \\ 
\toprule[1.2pt]

{\multirow{4}{*}{adult}} & {\multirow{4}{*}{(13}} & {\multirow{4}{*}{200)}} & {\multirow{4}{*}{81.37}} & {\multirow{4}{*}{6.8$\pm$ 1.2}} & 98 & 6.8$\pm$ 1.1 & 100$\pm$ 0.0 & 100 & 0.003 & 6.3$\pm$ 0.9 & 99.61$\pm$ 0.6 & 96 & 0.023 & 4.8$\pm$ 1.3 & 98.73$\pm$ 0.5 & 48 & 0.059 \\
  &   &   &   &   & 95 & 6.8$\pm$ 1.1 & 99.99$\pm$ 0.2 & 100 & 0.074 & 5.9$\pm$ 1.0 & 98.87$\pm$ 1.8 & 99 & 0.058 & 3.9$\pm$ 1.0 & 96.93$\pm$ 1.1 & 80 & 0.071 \\
  &   &   &   &   & 93 & 6.8$\pm$ 1.1 & 99.97$\pm$ 0.4 & 100 & 0.104 & 5.7$\pm$ 1.3 & 98.34$\pm$ 2.6 & 100 & 0.086 & 3.4$\pm$ 0.9 & 95.21$\pm$ 1.6 & 90 & 0.093 \\
  &   &   &   &   & 90 & 6.8$\pm$ 1.1 & 99.95$\pm$ 0.6 & 100 & 0.164 & 5.5$\pm$ 1.4 & 97.86$\pm$ 3.4 & 100 & 0.100 & 3.0$\pm$ 0.8 & 93.46$\pm$ 1.5 & 94 & 0.103 \\
\midrule
{\multirow{4}{*}{agaricus}} & {\multirow{4}{*}{(23}} & {\multirow{4}{*}{200)}} & {\multirow{4}{*}{95.41}} & {\multirow{4}{*}{10.3$\pm$ 2.5}} & 98 & 7.7$\pm$ 2.7 & 99.12$\pm$ 0.8 & 92 & 0.593 & 6.4$\pm$ 3.0 & 98.75$\pm$ 0.6 & 87 & 0.763 & 6.0$\pm$ 3.1 & 98.67$\pm$ 0.5 & 29 & 0.870 \\
  &   &   &   &   & 95 & 6.9$\pm$ 3.1 & 97.62$\pm$ 2.1 & 95 & 0.954 & 5.3$\pm$ 3.2 & 96.59$\pm$ 1.6 & 92 & 1.273 & 4.8$\pm$ 3.3 & 96.24$\pm$ 1.2 & 55 & 1.217 \\
  &   &   &   &   & 93 & 6.5$\pm$ 3.1 & 96.65$\pm$ 2.8 & 95 & 1.112 & 4.8$\pm$ 3.1 & 95.38$\pm$ 1.9 & 93 & 1.309 & 4.3$\pm$ 3.1 & 94.92$\pm$ 1.3 & 64 & 1.390 \\
  &   &   &   &   & 90 & 5.9$\pm$ 3.3 & 94.95$\pm$ 4.1 & 96 & 1.332 & 4.0$\pm$ 3.0 & 92.60$\pm$ 2.8 & 95 & 1.598 & 3.6$\pm$ 2.8 & 92.08$\pm$ 1.7 & 76 & 1.830 \\
\midrule
{\multirow{4}{*}{chess}} & {\multirow{4}{*}{(37}} & {\multirow{4}{*}{200)}} & {\multirow{4}{*}{88.34}} & {\multirow{4}{*}{12.1$\pm$ 3.7}} & 98 & 8.1$\pm$ 4.1 & 99.27$\pm$ 0.6 & 64 & 0.383 & 5.9$\pm$ 4.9 & 98.70$\pm$ 0.4 & 64 & 0.454 & 5.7$\pm$ 5.0 & 98.65$\pm$ 0.4 & 46 & 0.457 \\
  &   &   &   &   & 95 & 7.7$\pm$ 3.8 & 98.51$\pm$ 1.4 & 68 & 0.404 & 5.5$\pm$ 4.4 & 97.90$\pm$ 0.9 & 64 & 0.483 & 5.3$\pm$ 4.5 & 97.85$\pm$ 0.8 & 46 & 0.478 \\
  &   &   &   &   & 93 & 7.3$\pm$ 3.5 & 97.56$\pm$ 2.4 & 68 & 0.419 & 5.0$\pm$ 4.1 & 96.26$\pm$ 2.2 & 64 & 0.485 & 4.8$\pm$ 4.1 & 96.21$\pm$ 2.1 & 64 & 0.493 \\
  &   &   &   &   & 90 & 7.3$\pm$ 3.5 & 97.29$\pm$ 2.9 & 70 & 0.413 & 4.9$\pm$ 4.0 & 95.99$\pm$ 2.6 & 64 & 0.483 & 4.8$\pm$ 4.0 & 95.93$\pm$ 2.5 & 64 & 0.543 \\
\midrule
{\multirow{4}{*}{vote}} & {\multirow{4}{*}{(17}} & {\multirow{4}{*}{81)}} & {\multirow{4}{*}{89.66}} & {\multirow{4}{*}{5.3$\pm$ 1.4}} & 98 & 5.3$\pm$ 1.4 & 100$\pm$ 0.0 & 100 & 0.000 & 5.3$\pm$ 1.3 & 99.95$\pm$ 0.2 & 100 & 0.007 & 4.6$\pm$ 1.1 & 99.60$\pm$ 0.4 & 64 & 0.014 \\
  &   &   &   &   & 95 & 5.3$\pm$ 1.4 & 100$\pm$ 0.0 & 100 & 0.000 & 5.3$\pm$ 1.3 & 99.93$\pm$ 0.3 & 100 & 0.008 & 4.1$\pm$ 1.0 & 98.25$\pm$ 1.7 & 64 & 0.018 \\
  &   &   &   &   & 93 & 5.3$\pm$ 1.4 & 100$\pm$ 0.0 & 100 & 0.000 & 5.2$\pm$ 1.3 & 99.78$\pm$ 1.1 & 100 & 0.012 & 4.1$\pm$ 0.9 & 98.10$\pm$ 1.9 & 64 & 0.018 \\
  &   &   &   &   & 90 & 5.3$\pm$ 1.4 & 100$\pm$ 0.0 & 100 & 0.000 & 5.2$\pm$ 1.3 & 99.78$\pm$ 1.1 & 100 & 0.012 & 4.0$\pm$ 1.2 & 97.24$\pm$ 3.1 & 64 & 0.022 \\
\midrule
{\multirow{4}{*}{kr-vs-kp}} & {\multirow{4}{*}{(37}} & {\multirow{4}{*}{200)}} & {\multirow{4}{*}{88.07}} & {\multirow{4}{*}{12.2$\pm$ 3.9}} & 98 & 7.8$\pm$ 4.2 & 99.19$\pm$ 0.5 & 64 & 0.387 & 6.5$\pm$ 4.7 & 98.99$\pm$ 0.4 & 64 & 0.427 & 6.1$\pm$ 4.9 & 98.88$\pm$ 0.3 & 43 & 0.457 \\
  &   &   &   &   & 95 & 7.3$\pm$ 3.9 & 98.29$\pm$ 1.4 & 64 & 0.416 & 6.0$\pm$ 4.3 & 97.89$\pm$ 1.1 & 64 & 0.453 & 5.5$\pm$ 4.5 & 97.79$\pm$ 0.9 & 43 & 0.462 \\
  &   &   &   &   & 93 & 6.9$\pm$ 3.5 & 97.21$\pm$ 2.5 & 69 & 0.422 & 5.6$\pm$ 3.8 & 96.82$\pm$ 2.2 & 64 & 0.448 & 5.2$\pm$ 4.0 & 96.71$\pm$ 2.1 & 43 & 0.468 \\
  &   &   &   &   & 90 & 6.8$\pm$ 3.5 & 96.65$\pm$ 3.1 & 69 & 0.418 & 5.4$\pm$ 3.8 & 95.69$\pm$ 3.0 & 64 & 0.468 & 5.0$\pm$ 4.0 & 95.59$\pm$ 2.8 & 61 & 0.487 \\
\midrule
{\multirow{4}{*}{mushroom}} & {\multirow{4}{*}{(23}} & {\multirow{4}{*}{200)}} & {\multirow{4}{*}{95.51}} & {\multirow{4}{*}{10.7$\pm$ 2.3}} & 98 & 7.5$\pm$ 2.4 & 98.99$\pm$ 0.7 & 90 & 0.641 & 6.5$\pm$ 2.6 & 98.74$\pm$ 0.5 & 83 & 0.751 & 6.3$\pm$ 2.7 & 98.70$\pm$ 0.4 & 18 & 0.828 \\
  &   &   &   &   & 95 & 6.5$\pm$ 2.6 & 97.35$\pm$ 1.8 & 96 & 1.011 & 5.1$\pm$ 2.5 & 96.52$\pm$ 1.0 & 90 & 1.130 & 5.0$\pm$ 2.5 & 96.39$\pm$ 0.8 & 54 & 1.113 \\
  &   &   &   &   & 93 & 5.8$\pm$ 2.8 & 95.77$\pm$ 2.7 & 96 & 1.257 & 4.4$\pm$ 2.5 & 94.67$\pm$ 1.6 & 94 & 1.297 & 4.2$\pm$ 2.4 & 94.48$\pm$ 1.3 & 65 & 1.324 \\
  &   &   &   &   & 90 & 5.3$\pm$ 3.0 & 94.01$\pm$ 3.9 & 97 & 1.455 & 3.8$\pm$ 2.3 & 92.36$\pm$ 2.2 & 96 & 1.543 & 3.6$\pm$ 2.2 & 92.07$\pm$ 1.6 & 76 & 1.650 \\
\midrule
{\multirow{4}{*}{threeOf9}} & {\multirow{4}{*}{(10}} & {\multirow{4}{*}{103)}} & {\multirow{4}{*}{83.13}} & {\multirow{4}{*}{4.2$\pm$ 0.4}} & 98 & 4.2$\pm$ 0.4 & 100$\pm$ 0.0 & 100 & 0.000 & 4.2$\pm$ 0.4 & 100$\pm$ 0.0 & 100 & 0.000 & 4.2$\pm$ 0.4 & 100$\pm$ 0.0 & 78 & 0.001 \\
  &   &   &   &   & 95 & 4.2$\pm$ 0.4 & 100$\pm$ 0.0 & 100 & 0.000 & 4.2$\pm$ 0.4 & 100$\pm$ 0.0 & 100 & 0.000 & 4.0$\pm$ 0.2 & 99.23$\pm$ 1.4 & 100 & 0.002 \\
  &   &   &   &   & 93 & 4.2$\pm$ 0.4 & 100$\pm$ 0.0 & 100 & 0.000 & 4.2$\pm$ 0.4 & 100$\pm$ 0.0 & 100 & 0.000 & 3.9$\pm$ 0.2 & 99.20$\pm$ 1.5 & 100 & 0.002 \\
  &   &   &   &   & 90 & 4.2$\pm$ 0.4 & 100$\pm$ 0.0 & 100 & 0.000 & 4.2$\pm$ 0.4 & 100$\pm$ 0.0 & 100 & 0.000 & 3.8$\pm$ 0.4 & 98.29$\pm$ 3.3 & 100 & 0.003 \\
\midrule
{\multirow{4}{*}{xd6}} & {\multirow{4}{*}{(10}} & {\multirow{4}{*}{176)}} & {\multirow{4}{*}{81.36}} & {\multirow{4}{*}{4.5$\pm$ 0.9}} & 98 & 4.5$\pm$ 0.8 & 100$\pm$ 0.0 & 100 & 0.000 & 4.5$\pm$ 0.8 & 100$\pm$ 0.0 & 100 & 0.000 & 4.5$\pm$ 0.8 & 100$\pm$ 0.0 & 73 & 0.001 \\
  &   &   &   &   & 95 & 4.5$\pm$ 0.8 & 100$\pm$ 0.0 & 100 & 0.000 & 4.5$\pm$ 0.8 & 100$\pm$ 0.0 & 100 & 0.000 & 4.5$\pm$ 0.8 & 100$\pm$ 0.0 & 73 & 0.001 \\
  &   &   &   &   & 93 & 4.5$\pm$ 0.8 & 100$\pm$ 0.0 & 100 & 0.000 & 4.5$\pm$ 0.8 & 100$\pm$ 0.0 & 100 & 0.000 & 4.3$\pm$ 0.4 & 98.30$\pm$ 2.7 & 73 & 0.001 \\
  &   &   &   &   & 90 & 4.5$\pm$ 0.8 & 100$\pm$ 0.0 & 100 & 0.000 & 4.5$\pm$ 0.8 & 100$\pm$ 0.0 & 100 & 0.000 & 4.3$\pm$ 0.4 & 98.30$\pm$ 2.7 & 73 & 0.002 \\
\midrule
{\multirow{4}{*}{mamo}} & {\multirow{4}{*}{(14}} & {\multirow{4}{*}{53)}} & {\multirow{4}{*}{80.21}} & {\multirow{4}{*}{4.9$\pm$ 0.8}} & 98 & 4.9$\pm$ 0.7 & 100$\pm$ 0.0 & 100 & 0.000 & 4.9$\pm$ 0.7 & 100$\pm$ 0.0 & 100 & 0.000 & 4.6$\pm$ 0.6 & 99.66$\pm$ 0.5 & 53 & 0.007 \\
  &   &   &   &   & 95 & 4.9$\pm$ 0.7 & 100$\pm$ 0.0 & 100 & 0.000 & 4.9$\pm$ 0.7 & 100$\pm$ 0.0 & 100 & 0.000 & 3.9$\pm$ 0.6 & 97.80$\pm$ 1.6 & 85 & 0.009 \\
  &   &   &   &   & 93 & 4.9$\pm$ 0.7 & 100$\pm$ 0.0 & 100 & 0.000 & 4.9$\pm$ 0.7 & 100$\pm$ 0.0 & 100 & 0.000 & 3.9$\pm$ 0.6 & 97.68$\pm$ 1.7 & 85 & 0.009 \\
  &   &   &   &   & 90 & 4.9$\pm$ 0.7 & 100$\pm$ 0.0 & 100 & 0.000 & 4.9$\pm$ 0.7 & 100$\pm$ 0.0 & 100 & 0.000 & 3.6$\pm$ 0.8 & 96.18$\pm$ 3.2 & 96 & 0.011 \\
\midrule
{\multirow{4}{*}{tumor}} & {\multirow{4}{*}{(16}} & {\multirow{4}{*}{104)}} & {\multirow{4}{*}{83.21}} & {\multirow{4}{*}{5.3$\pm$ 0.9}} & 98 & 5.3$\pm$ 0.8 & 100$\pm$ 0.0 & 100 & 0.000 & 5.2$\pm$ 0.7 & 99.96$\pm$ 0.2 & 100 & 0.008 & 4.1$\pm$ 0.7 & 99.41$\pm$ 0.5 & 91 & 0.012 \\
  &   &   &   &   & 95 & 5.3$\pm$ 0.8 & 100$\pm$ 0.0 & 100 & 0.000 & 5.2$\pm$ 0.6 & 99.83$\pm$ 0.7 & 100 & 0.012 & 3.2$\pm$ 0.6 & 96.02$\pm$ 1.5 & 94 & 0.016 \\
  &   &   &   &   & 93 & 5.3$\pm$ 0.8 & 100$\pm$ 0.0 & 100 & 0.000 & 5.2$\pm$ 0.6 & 99.74$\pm$ 1.2 & 100 & 0.014 & 3.1$\pm$ 0.7 & 95.50$\pm$ 1.4 & 95 & 0.016 \\
  &   &   &   &   & 90 & 5.3$\pm$ 0.8 & 100$\pm$ 0.0 & 100 & 0.000 & 5.1$\pm$ 0.7 & 99.67$\pm$ 1.4 & 100 & 0.016 & 3.0$\pm$ 0.6 & 95.30$\pm$ 1.6 & 95 & 0.017 \\

\bottomrule[1.2pt]
\end{tabular}
}
\caption{ Assessing ApproxPAXp explanations for NBCs. 
	Columns {\bf \#F} and  {\bf \#I} show, respectively, number of features and 
	tested instances in the Dataset.
	 Column {\bf A\%} reports in (\%) the training accuracy of the classifier. 
	 Column {$\delta$} reports in (\%) the used value of the parameter $\delta$.
	 {\bf ApproxPAXp$_{\le 9}$}, {\bf ApproxPAXp$_{\le 7}$} and {\bf ApproxPAXp$_{\le 4}$} denote, 
	 respectively,  ApproxPAXp's of (target) length 9, 7 and 4.
	 Columns {\bf Length} and {\bf Precision} report, respectively, the average  
	 explanation length and the average explanation precision ($\pm$ denotes 
	 the standard  deviation). 
	 {\bf W\%} shows in (\%) the number of success/wins where the explanation 
	 size is less or equal than the target size. 
	 Finally, the average runtime  to compute an explanation is shown 
	 (in seconds) in {\bf Time}.
	 (Note that the reported average time is the mean of runtimes for instances 
	 we effectively computed an approximate explanation, namely instances that 
	 have AXp's of length longer than the target length; whereas for the remaining 
	 instances  the trimming process is skipped and the runtime is 0 sec, thus 
	 we exclude them when calculating the average.)
	 } 
\label{tab:paxp-res}
\end{table*}

\section{Experimental Results} \label{sec:res}
This section evaluates the algorithm proposed for computing 
ApproxPAXp's.
The evaluation  aims at assessing not only the succinctness 
and precision of computed explanations but also the scalability 
of our solution.

\subsection{Experimental setup}
The benchmarks used in the experiments comprise publicly 
available and widely used datasets that originate from 
UCI ML Repository \cite{uci} and Penn ML Benchmarks \cite{pennml}.
The number of training data (resp.\ features) in the target
datasets varies from 336 to 14113 (resp.\ 10 to 37) and on average 
is 3999.1 (resp.\  20.0).
All the NBCs are trained using the learning tool
\emph{scikit-learn}~\cite{sklearn}.
The data split for  training and test data is set to 80\% and 
20\%, respectively.
Model accuracies are above 80\% for the training accuracy and 
above 75\% for the test accuracy.

A prototype implementation\footnote{All sources will be publicly 
available after that the paper gets accepted in a conference.} 
of the proposed
approach for computing relevant sets is developed in Python. 
To compute AXp's, we use the Perl script implemented by 
\cite{msgcin-nips20}\footnote{Publicly available from:
  \url{https://github.com/jpmarquessilva/expxlc}}.
The prototype implementation was tested with varying  the
threshold  $\delta \in \{0.90, 0.93, 0.95,  0.98\}$.
When converting probabilities from real values to integer values, the
selected number of decimal places is 3. 
(As outlined earlier, we observed that there is a negligible accuracy 
loss from using three decimal places.)
In order to produce explanations of size admissible
for the cognitive capacity of human decision makers~\cite{miller-pr56}, 
we selected three different target sizes for the explanations 
to compute: 9, 7 and 4, and we compute a ApproxPAXp for the input 
instance when its AXp $\fml{X}$ is larger than the target size 
(recall that $\fml{S}$ is initialized to $\fml{X}$); otherwise 
we consider the AXp is succinct and the explainer returns $\fml{X}$.
For example, assume the target size is 7, an instance $\mbf{v}_1$ with 
an AXp $\fml{S}_1$ of 5 features and an second instance $\mbf{v}_2$ with 
an AXp $\fml{S}_2$  of 8 features,  then for $\mbf{v}_1$ the output will be 
$\fml{S}_1$ and for $\mbf{v}_2$ the output will be a subset of $\fml{S}_2$.

For each dataset, we run the explainer on 200 instances randomly 
picked from the test data or on all  instances if there are less than
200. 

The experiments are conducted on a MacBook Air with a 1.1GHz
Quad-Core Intel Core~i5  CPU with 16 GByte RAM running 
macOS  Monterey.

\subsection{Results}
\cref{tab:paxp-res} summarizes the results of our experiments. 
For all tested values for the parameter threshold $\delta$ and 
target size, the reported results show the sizes and precisions 
of the computed explanations.
As can be observed for all considered settings, the approximate 
explanations are succinct, in particular the average 
sizes of the explanations are invariably lower than the target sizes.
Moreover, theses explanations  offer strong guarantees of precision, as 
their average precisions are strictly greater than $\delta$ with significant gaps 
(e.g.\  above 97\%, in column ApproxPAXp$_{\le 7}$, for datasets \emph{adult}, 
\emph{vote}, \emph{threeOf9}, \emph{xd6}, \emph{mamo} and 
\emph{tumor} and above 95\% for \emph{chess} and \emph{kr-vs-kp}). 
%
An important observation from the results, is the gain of succinctness 
(explanation size) when comparing AXp's with ApproxPAXp's. Indeed, 
for some datasets, the AXp's are too large (e.g. for \emph{chess} and 
\emph{kr-vs-kp} datasets, the average number of features in the AXp's 
is 12),  exceeding the cognitive limits of human decision 
makers~\cite{miller-pr56} (limited to 7 $\pm$ 2 features).   
To illustrate that, one can focus on the dataset  {\it agaricus} or {\it mushroom} 
and see that for a target size equals to 7 and $\delta = 0.95$, the average 
length of the ApproxPAXp's (i.e.\ 5.3 and 5.1, resp.) is 2 times less than 
the average length of the AXp's (i.e.\ 10.3 and 10.7, resp.).
Besides, the results show that $\delta = 0.95$ is a good probability threshold 
to guarantee highly precise and short approximate explanations.

Despite the complexity of the proposed approach being in pseudo polynomial, 
the results demonstrate that in practice the algorithm is effective and scales 
for large datasets.
As can be seen, the runtimes are negligible for all datasets, never exceeding 
2 seconds for the largest datasets (i.e. \emph{agaricus} or \emph{mushroom}) 
and the average is 0.33 seconds for all tested instances across all datasets and 
all settings.    
Furthermore, we point out that the implemented prototype was tested with 4 
decimal  places to assess further the scalability of the algorithm on larger 
DP tables, and the results show that computing  ApproxPAXp's is still 
feasible,  e.g.\  with {\it agaricus}  the average runtime for $\delta$ set to 
0.95 and target size to 7 is 10.08 seconds and 7.22 seconds for $\delta=0.98$. 

%
The table also reports the number of explanations being shorter than or of size 
equal to the target size over the total number of tested instances.  We observe 
that for both settings ApproxPAXp$_{\le 9}$ and ApproxPAXp$_{\le 7}$ and 
for the majority of datasets and with a few exceptions the fraction is significantly 
high, e.g.\  varying for 96\% to 100\% for \emph{adult} dataset. However, 
for ApproxPAXp$_{\le 4}$  despite the  poor percentage of wins for some 
datasets, it is the case that the average lengths of computed explanations 
are close to 4.   

Overall, the experiments demonstrate that our approach efficiently computes 
succinct and provably precise explanations for NBCs. 
The results also showcase empirically the advantage of the algorithm, i.e.\ in practice 
one may rely on the computation of ApproxPAXp’s, which pays off in terms of 
(1) performance, (2) succinctness and (3) sufficiently high probabilistic guarantees 
of precision.

\section{Conclusion} \label{sec:conc}

This paper builds on recent work on computing rigorous probabilistic
explanations~\cite{kutyniok-jair21}, and investigates the concrete
case of NBCs.
The paper proposes a pseudo-polynomial algorithm for computing the
number of points in feature space predicting a specific class, and
relates this problem with that of computing a rigorous probabilistic
explanation.
Furthermore, the paper proposes an algorithm for computing approximate
probabilistic explanations, which offers strong guarantees in terms of
precision.
The experimental results confirm that short and precise probabilistic
explanations can be efficiently computed in the case of NBCs.

Two lines of future work can be envisioned.  One line is to investigate  
the complexity of explaining multi-class NBCs and extend 
the approach for computing  approximate probabilistic explanations 
for multi-class Naive Bayes models. 
Furthermore, one might be interested in computing smallest  probabilistic 
explanations instead of approximates. Hence, another line of research
is to devise a logical (Satisfiability Modulo Theories, SMT) encoding
for computing cardinality minimal probabilistic explanations.

\subsection*{Acknowledgments}
  %
  This work was supported by the AI Interdisciplinary Institute ANITI, 
  funded by the French program ``Investing for the Future -- PIA3''
  under Grant agreement no.\ ANR-19-PI3A-0004, and by the H2020-ICT38
  project COALA ``Cognitive Assisted agile manufacturing for a Labor
  force supported by trustworthy Artificial intelligence''.
  %



\newtoggle{mkbbl}

\settoggle{mkbbl}{false}

\iftoggle{mkbbl}{
\bibliographystyle{ACM-Reference-Format}
\bibliography{refs,team,nips20}
}{

}

\end{document}